\documentclass[runningheads]{llncs}

\makeatletter
\g@addto@macro\normalsize{%
  \setlength\abovedisplayskip{4pt plus 2pt minus 2pt}%
  \setlength\belowdisplayskip{4pt plus 2pt minus 2pt}%
  \setlength\abovedisplayshortskip{2pt plus 2pt minus 1pt}%
  \setlength\belowdisplayshortskip{2pt plus 2pt minus 1pt}%
}
\makeatother

\usepackage{algorithm}
\usepackage{algorithmic}

\usepackage{amsmath,amssymb}

\usepackage{cite}
\usepackage{url}
\usepackage[hidelinks]{hyperref}

\usepackage{xcolor}
\usepackage{microtype}

\usepackage{booktabs}
\usepackage{multirow}
\usepackage{array}
\usepackage{makecell}
\usepackage{tabularx}

\usepackage[T1]{fontenc}
\usepackage{graphicx}
\begin{document}
\title{MoSSGate: Memory-Modulated State-Space Gating for Skin Lesion Segmentation}\titlerunning{MoSSGate}
\author{Anum Awan\inst{1}\orcidID{0090-0028-2604-2680} \and
Mahnoor Buriro\inst{1}\orcidID{0009-0004-1690-5063} \and
Muhammad Younas Khan\inst{1}\orcidID{0009-0002-1817-4693} \and
Md Imam Ahasan\inst{1}\orcidID{0009-0009-8407-2071}}
\authorrunning{Anum et al.}
%
\institute{College of Computer Science, Chongqing University, Chongqing, China \\
\email{anum23215@gmail.com}
}
\maketitle              
\begin{abstract}
Accurate skin lesion segmentation is crucial for reliable computer-aided dermatological diagnosis, yet existing convolutional and transformer-based models often struggle to jointly capture long-range spatial dependencies and fine boundary details under limited computational budgets. This trade-off between global context modeling and boundary-aware localization frequently leads to over-segmentation, fragmented predictions, or missing thin peripheral structures. To address this challenge, we propose MoSSGate, a plug-and-play module for U-Net that integrates (i) boundary-aware spatial gating to restrict long-range propagation to informative regions, (ii) an external memory modulator that provides sample-adaptive dynamic control, and (iii) parallel 2D state-space modeling for efficient global context aggregation with linear complexity. The proposed design enables adaptive, context-aware information propagation while preserving sharp and accurate lesion boundaries. Extensive experiments on the ISIC 2017 and ISIC 2018 benchmarks demonstrate state-of-the-art accuracy with strong efficiency, achieving 86.3\% and 85.9\% mIoU and 92.6\% and 90.6\% Dice, respectively, while requiring substantially fewer FLOPs than most competing CNN-based methods. These results highlight a favorable accuracy efficiency trade-off for high-resolution medical image segmentation.

\end{abstract}
\section{Introduction}

Accurate medical image segmentation is essential for computer-aided diagnosis and treatment planning. In dermoscopy, precise delineation of skin lesions is crucial for early melanoma detection, where small boundary deviations may lead to significant clinical errors. Consequently, automated skin lesion segmentation remains an active research topic~\cite{wu2024mhorunet, wu2024hsh}. Deep learning has driven major progress in this area. Convolutional neural networks (CNNs) effectively model local appearance but are limited in capturing long-range dependencies and global shape. Vision transformers (ViTs)~\cite{badar2025transformer} provide global context via self-attention, yet often sacrifice fine local detail and incur high computational cost due to quadratic complexity. Balancing global reasoning, local precision, and efficiency is therefore a central challenge, especially for high-resolution medical images.

U-Net~\cite{ronneberger2015u} and its encoder–decoder variants dominate medical segmentation thanks to multi-scale skip-connected feature fusion. Numerous extensions enhance this backbone using dense connections, attention, or transformer modules~\cite{chen2021transunet, oktay2018attention, wu2022fat}, and MLP-style blocks~\cite{yu2022s2}. However, two difficulties persist in skin lesion segmentation. First, high-dimensional features contain substantial channel redundancy, weakening discriminative localization, prior work promotes channel diversity via differential or reorganization operations~\cite{zhao2023m, qiu2021slimconv}. Second, dermoscopic images exhibit low-contrast, irregular boundaries and strong appearance variability, requiring both global context and sharp local boundary modeling. Multi-scale context aggregation and boundary-aware supervision partially mitigate these issues~\cite{zhao2023mms, he2023h2former, ruan2023ege, lin2023rethinking} but remain fragile in complex cases. State space models (SSMs) provide an efficient alternative for long-range modeling with linear complexity~\cite{mehta2022long, wang2023selective}. Structured SSMs such as S4~\cite{gu2021efficiently} and its selective variant Mamba~\cite{gu2024mamba}, together with 2D visual adaptations based on cross-directional scans~\cite{liu2024vmamba}, enable global spatial interaction at low computational cost. This makes SSMs particularly appealing for medical segmentation, where global coherence and boundary fidelity must be achieved under tight efficiency constraints.

\begin{figure}[t]
    \centering
    \includegraphics[width=.75\linewidth]{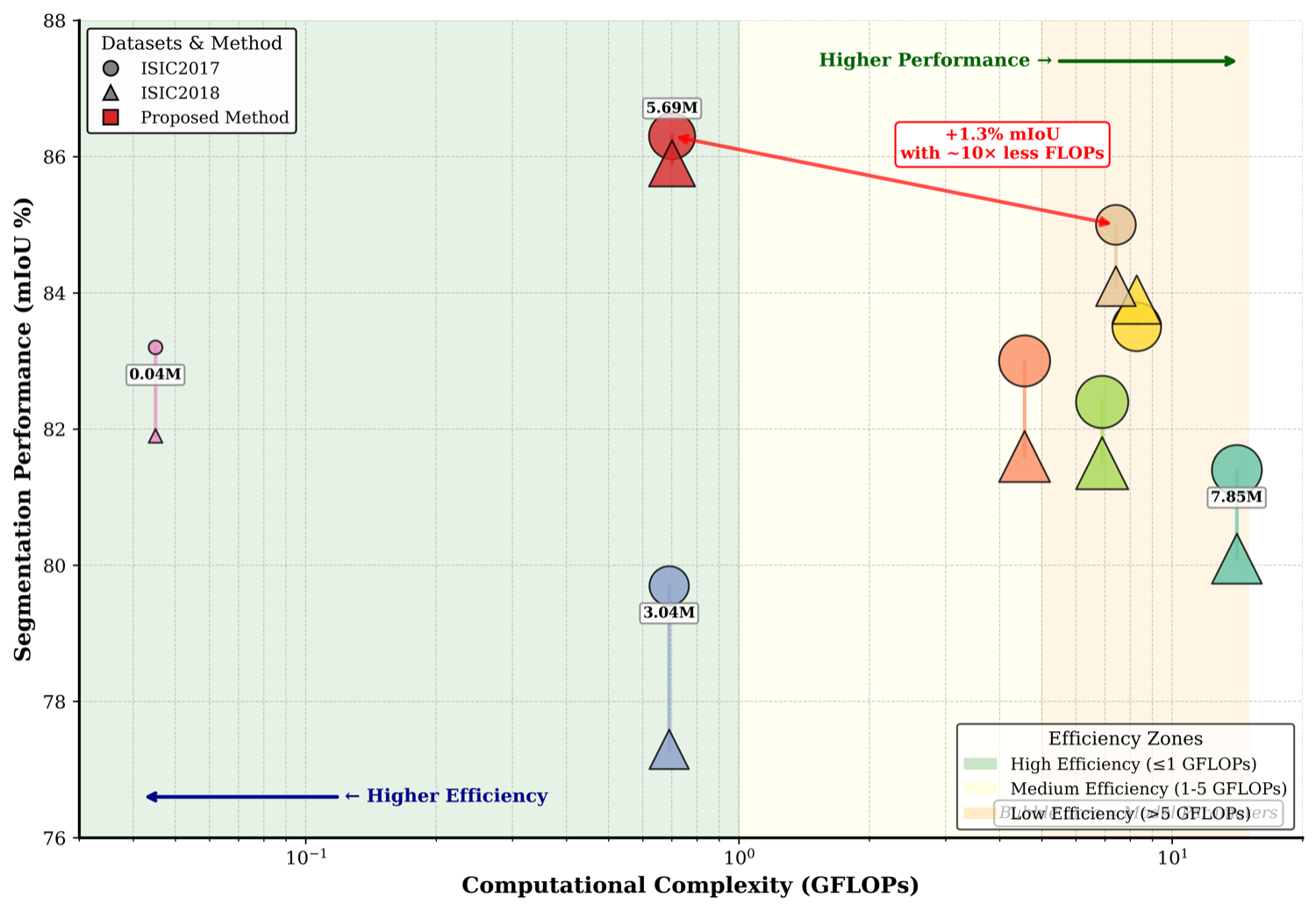}
    \caption{Efficiency-performance trade-off on ISIC 2017 and ISIC 2018. Bubble size denotes parameter count, and the x-axis (log scale) denotes GFLOPs. Our method achieves the highest mIoU with low computational cost, indicating a strong accuracy-efficiency trade-off.}
    \label{fig:efficiency_performance_tradeoff}
\end{figure}

We propose MoSSGate, a memory-modulated state-space gating module for U-Net. It combines efficient long-range spatial modeling with boundary-aware spatial gating and sample-adaptive contextual modulation. By selectively routing informative regions into parallel 2D state-space recurrences while suppressing irrelevant background, MoSSGate achieves accurate and robust lesion delineation with minimal overhead, as illustrated by the efficiency-performance comparison in Fig. \ref{fig:efficiency_performance_tradeoff}. Our main contributions are:
\begin{enumerate}
    \item A parallel 2D state-space module integrated into U-Net, whose dynamics are modulated by a lightweight external memory for sample-adaptive long-range context modeling with linear complexity.
    \item A learnable spatial gate that highlights probable lesion regions before long-range propagation, improving boundary adherence and reducing background interference.
    \item A plug-in design for deep U-Net stages that achieves state-of-the-art results on ISIC benchmarks with substantially fewer FLOPs than competing CNN and transformer-based methods.
\end{enumerate}

\section{Related Work}

\subsection{Skin Lesion Segmentation}
Early deep learning methods for skin lesion segmentation are mainly based on CNN encoder-decoder architectures such as U-Net and its variants~\cite{ronneberger2015u, li2025edgrnet, GeGLUNet}, which learn strong local features but have limited ability to model global context. To address this, hybrid CNN-Transformer models introduce self-attention for long-range dependency modeling and richer contextual interaction~\cite{ashish2017attention, chen2023transattunet, khan2025medfusion}. Subsequent U-shaped networks incorporate attention and transformer mechanisms to emphasize salient regions and capture non-local relations~\cite{wang2022uctransnet, abbas2025dualattendmed}. Another line of work explicitly enhances boundary quality using shape or contour priors, including star-shaped constraints~\cite{huang2025lmfa}, edge/boundary attention modules~\cite{wang2022eanet}, and supervision from representative boundary points~\cite{lee2020structure, wang2021boundary}. In parallel, multi-scale and parallel feature extraction strategies improve robustness to artifacts and appearance variability~\cite{zhao2023mms, li2025vmc, he2023h2former, abbas2025gradient}, while channel diversification methods reduce feature redundancy~\cite{zhao2023m, qiu2021slimconv}. Although effective in isolation, these approaches struggle to simultaneously achieve efficient global modeling, precise boundary localization, and redundancy suppression.

Recently, state-space models (SSMs) have been introduced to vision as an efficient alternative to self-attention for long-range dependency modeling. Structured SSMs such as S4 and Mamba~\cite{gu2021efficiently, gu2024mamba}, together with visual adaptations based on cross-directional scanning~\cite{liu2024vmamba}, enable global spatial interaction with linear complexity. However, existing visual SSMs mainly target general recognition or dense prediction and do not explicitly address boundary sensitivity, feature redundancy, or sample-adaptive contextual modulation required by challenging medical segmentation tasks.

\subsection{Skip Connections in U-Shaped Networks}

Skip connections restore spatial detail by fusing high-resolution encoder features into the decoder~\cite{ronneberger2015u}. Dense variants such as U-Net++ increase cross-scale information flow but also introduce substantial redundancy. Later work refines skip fusion using re-weighting, normalization, and boundary-aware processing~\cite{ruan2023ege, lin2023rethinking, ruan2022malunet}. Attention and transformer-based skip pathways further improve semantic alignment between encoder and decoder features~\cite{ wu2022fat}, while channel or spatial cross-attention reduces cross-level semantic gaps~\cite{wang2022uctransnet, ates2023dual}. However, increasingly complex attention mechanisms may propagate irrelevant correlations and amplify noise. Rather than complicating skip fusion, our method enhances deep encoder features using efficient state-space modeling with spatial gating and memory-based modulation before standard U-Net skips.
\begin{figure*}[!ht]
    \centering
    \includegraphics[width=\linewidth]{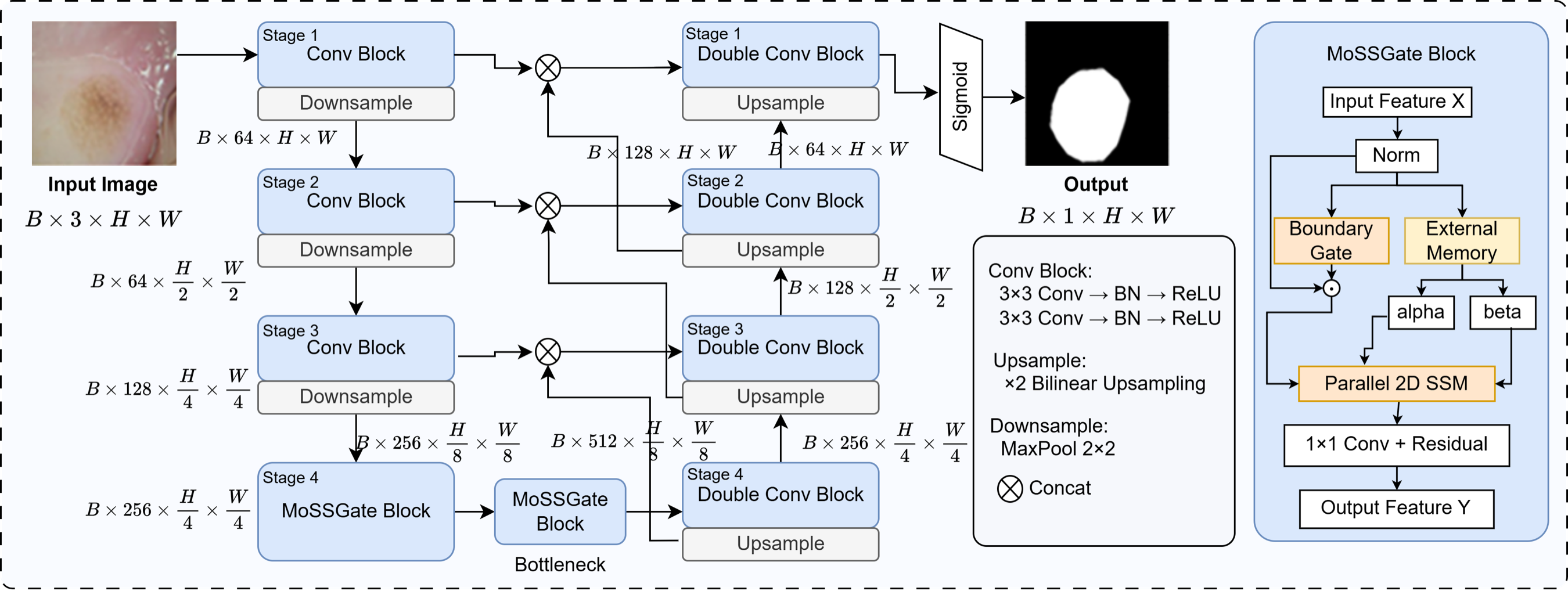}
    \caption{Overall architecture of the proposed U-Net with MoSSGate. Standard U-Net encoding–decoding with skip concatenation is preserved, while deep encoder and bottleneck blocks are replaced by MoSSGate for boundary-aware gating and efficient long-range feature modeling.}
    \label{fig:overall}
\end{figure*}
\vspace{-16pt}
\section{Method}
\label{sec:method}

\subsection{Overall Network Architecture}
\label{sec:overall}

A standard U-Net encoder-decoder is adopted as the backbone. Let an input image be denoted by
$I \in \mathbb{R}^{B \times 3 \times H_0 \times W_0}$, where $B$ is the batch size and $(H_0, W_0)$ is the input spatial resolution. The encoder produces a hierarchy of feature maps
$\{X^{(s)}\}_{s=0}^{S}$, where $X^{(s)} \in \mathbb{R}^{B \times C_s \times H_s \times W_s}$ and $(H_s, W_s)$ decreases with stage index $s$ through $2\times$ downsampling. The decoder reconstructs a dense prediction by progressive upsampling and concatenation with encoder skip features. The proposed design preserves the U-Net macro-structure (number of stages, skip connections, and resolution schedule) and substitutes selected standard convolutional blocks in deep encoder stages and at the bottleneck with \emph{MoSSGate} blocks. In the implementation, the substitution is parameterized by stage identifiers and is typically applied at the bottleneck stage (denoted ``enc4'') and optionally at the deepest encoder stage preceding the bottleneck (``enc3''), where the spatial resolution is lowest and long-range interactions have reduced computational cost. The network outputs a single-channel logit map
$P \in \mathbb{R}^{B \times 1 \times H_0 \times W_0}$, which is converted to a probability map by sigmoid during inference. An overview of the proposed architecture is illustrated in Fig.~\ref{fig:overall}.

\subsection{Boundary Gate for Spatial Control}
\label{sec:boundary_gate}
Given an input feature map $X \in \mathbb{R}^{B \times C \times H \times W}$ at a MoSSGate location, a spatial gate $G \in \mathbb{R}^{B \times 1 \times H \times W}$ is computed to modulate the input drive to long-range modeling. The gate is constructed from depthwise dilated convolutions followed by a pointwise projection:
\begin{equation}
\label{eq:gate_raw}
U = \mathrm{DWConv}_{r=1}(X) + \mathrm{DWConv}_{r=d}(X),
\end{equation}
\begin{equation}
\label{eq:gate}
G = \sigma\!\left(\mathrm{PW}(U)\right),
\end{equation}
where $\mathrm{DWConv}_{r}(\cdot)$ denotes a depthwise $3 \times 3$ convolution with dilation rate $r$ and channel grouping equal to $C$, $\mathrm{PW}(\cdot)$ is a $1\times1$ convolution mapping from $C$ channels to $1$ channel, and $\sigma(\cdot)$ is the sigmoid function. Eq.~\eqref{eq:gate_raw} aggregates local responses at two receptive-field scales, while Eq.~\eqref{eq:gate} produces a single-channel control map shared across channels. The gated feature map $\tilde{X} \in \mathbb{R}^{B \times C \times H \times W}$ is defined by elementwise modulation:
\begin{equation}
\label{eq:gated_input}
\tilde{X} = X \odot G,
\end{equation}
where $\odot$ denotes broadcasted elementwise multiplication between $X$ and $G$ over the channel dimension. Eq.~\eqref{eq:gated_input} suppresses spatial locations with low gate responses prior to long-range propagation, thereby restricting the SSM input drive to regions indicated by the learned spatial control.

\subsection{External Memory Modulator}
\label{sec:memory_modulator}

A lightweight external memory is used to generate sample-conditioned modulation parameters for the state-space operator. Let $X \in \mathbb{R}^{B \times C \times H \times W}$ be the (normalized) input feature map to MoSSGate. A global feature statistic $z \in \mathbb{R}^{B \times C}$ is computed by global average pooling:
\begin{equation}
\label{eq:gap}
z = \mathrm{GAP}(X), \qquad z_{b,c} = \frac{1}{HW}\sum_{i=1}^{H}\sum_{j=1}^{W} X_{b,c,i,j}.
\end{equation}
The memory consists of learnable keys and values, $M_K, M_V \in \mathbb{R}^{m \times C}$, where $m$ is the number of memory slots. Memory retrieval is implemented by attention over memory slots:
\begin{equation}
\label{eq:mem_attn}
A = \mathrm{softmax}(z M_K^\top), \qquad A \in \mathbb{R}^{B \times m},
\end{equation}
\begin{equation}
\label{eq:mem_ctx}
c = A M_V, \qquad c \in \mathbb{R}^{B \times C}.
\end{equation}
Eq.~\eqref{eq:mem_attn} computes per-sample attention weights across memory slots, and Eq.~\eqref{eq:mem_ctx} produces a retrieved context vector $c$.
Two modulation vectors $\alpha, \beta \in \mathbb{R}^{B \times C}$ are generated by linear projections followed by bounded nonlinearity:
\begin{equation}
\label{eq:alpha_beta}
\alpha = \tanh(W_\alpha c + b_\alpha), \qquad
\beta  = \tanh(W_\beta  c + b_\beta),
\end{equation}
where $W_\alpha, W_\beta \in \mathbb{R}^{C \times C}$ and $b_\alpha, b_\beta \in \mathbb{R}^{C}$ are learned parameters. Eq.~\eqref{eq:alpha_beta} constrains the modulation to $[-1,1]$ elementwise. In the implementation, $W_\alpha$ and $W_\beta$ are initialized to zero to start from an unmodulated baseline and progressively learn sample-conditioned modulation during training.

\begin{algorithm}[!ht]
\caption{MoSSGate: Memory-Modulated 2D State-Space Gated Block}
\label{alg:mossgate}
\begin{algorithmic}[1]
\REQUIRE Input feature map $X$, groups $K$, memory size $m$, dilation $d$
\ENSURE Output feature map $Y$

\STATE \textbf{Boundary-aware gating}
\STATE $\bar{X} \leftarrow \mathrm{Norm}(X)$
\STATE $G \leftarrow \sigma\!\big(\mathrm{PW}(\mathrm{DWConv}_{1}(\bar{X}) + \mathrm{DWConv}_{d}(\bar{X}))\big)$
\STATE $\tilde{X} \leftarrow \bar{X} \odot G$

\STATE \textbf{Memory-based modulation}
\STATE $z \leftarrow \mathrm{GAP}(\bar{X})$
\STATE $c \leftarrow \mathrm{softmax}(z M_K^\top) M_V$
\STATE $\alpha \leftarrow \tanh(W_\alpha c + b_\alpha),\;
       \beta \leftarrow \tanh(W_\beta c + b_\beta)$

\STATE \textbf{Group-wise 2D state-space propagation}
\STATE Split $\tilde{X},\alpha,\beta$ into $\{\tilde{X}^{(k)},\alpha^{(k)},\beta^{(k)}\}_{k=1}^K$
\FOR{$k = 1$ \TO $K$}
    \STATE $a^{(k)} \leftarrow 
    \mathrm{clip}\!\big(\sigma(\ell^{(k)}) \odot (1+\alpha^{(k)}),\,0,\,0.999\big)$
    \STATE $O^{(k)} \leftarrow \mathrm{Avg}_{\delta \in \{\rightarrow,\leftarrow,\downarrow,\uparrow\}}
           \mathrm{SSMScan}(\tilde{X}^{(k)}, a^{(k)}, \delta)$
    \STATE $O^{(k)} \leftarrow 
    O^{(k)} + d_s^{(k)} \odot (1+\beta^{(k)}) \odot \tilde{X}^{(k)}$
\ENDFOR

\STATE \textbf{Fusion and residual output}
\STATE $O \leftarrow \mathrm{Concat}(O^{(1)},\ldots,O^{(K)})$
\STATE $Y \leftarrow X + \delta\!\big(\mathrm{Norm}(\mathrm{Proj}(O))\big)$
\RETURN $Y$
\end{algorithmic}
\end{algorithm}

\subsection{Parallel 2D State-Space Modeling}
\label{sec:ssm2d}
MoSSGate applies parallel state-space modeling to the gated feature map $\tilde{X}$ in Eq.~\eqref{eq:gated_input}. Channels are partitioned into $K$ groups, with $C_k = C/K$ channels per group:
\begin{equation}
\label{eq:split}
\tilde{X} = \mathrm{Concat}\!\left(\tilde{X}^{(1)}, \ldots, \tilde{X}^{(K)}\right), \qquad
\tilde{X}^{(k)} \in \mathbb{R}^{B \times C_k \times H \times W}.
\end{equation}
The modulation vectors $\alpha$ and $\beta$ are partitioned analogously, yielding $\alpha^{(k)}, \beta^{(k)} \in \mathbb{R}^{B \times C_k}$. For each group $k$, a 2D scan is implemented as an average over four directional 1D recurrences. Let $u^{(k)} \in \mathbb{R}^{B \times C_k \times L}$ be a 1D sequence obtained by reshaping $\tilde{X}^{(k)}$ along a chosen scan direction, with $L$ equal to the scan length (either $W$ for row-wise scans or $H$ for column-wise scans, repeated across rows/columns by reshaping). A stable per-channel recurrence is defined as
\begin{equation}
\label{eq:ssm_rec}
y^{(k)}_t = a^{(k)} \odot y^{(k)}_{t-1} + \left(1 - a^{(k)}\right)\odot u^{(k)}_t, \qquad t = 1,\ldots,L,
\end{equation}
where $y^{(k)}_t \in \mathbb{R}^{B \times C_k}$ is the state output at step $t$ and $a^{(k)} \in (0,1)^{B \times C_k}$ is a per-sample, per-channel decay factor. Eq.~\eqref{eq:ssm_rec} corresponds to exponential smoothing with channelwise decay and has linear complexity in $L$. The decay factor is obtained from a learned base parameter and the memory-derived modulation $\alpha^{(k)}$. Let $\ell^{(k)} \in \mathbb{R}^{C_k}$ be a learnable logit parameter. The base decay is $a_0^{(k)} = \sigma(\ell^{(k)}) \in (0,1)^{C_k}$. The modulated decay is then defined by
\begin{equation}
\label{eq:mod_decay}
a^{(k)} = \mathrm{clip}\!\left(a_0^{(k)} \odot \left(1 + \alpha^{(k)}\right),\, 0,\, 0.999\right),
\end{equation}
where $\mathrm{clip}(\cdot)$ enforces numerical stability. Eq.~\eqref{eq:mod_decay} introduces sample-conditioned control of the recurrence dynamics through $\alpha^{(k)}$. To preserve local evidence, a modulated skip term is added at each scan step. Let $d^{(k)} \in \mathbb{R}^{C_k}$ be a learnable per-channel scale. The step output is defined as
\begin{equation}
\label{eq:skip}
o^{(k)}_t = y^{(k)}_t + d^{(k)} \odot \left(1 + \beta^{(k)}\right)\odot u^{(k)}_t.
\end{equation}
Eq.~\eqref{eq:skip} modulates the skip contribution using $\beta^{(k)}$ while retaining channelwise scaling via $d^{(k)}$. Directional scanning is applied along four directions: left-to-right, right-to-left, top-to-bottom, and bottom-to-top. Each direction produces a tensor in $\mathbb{R}^{B \times C_k \times H \times W}$ after inverse reshaping, the group output is computed by averaging:
\begin{equation}
\label{eq:dir_avg}
O^{(k)} = \frac{1}{4}\left(O^{(k)}_{\rightarrow} + O^{(k)}_{\leftarrow} + O^{(k)}_{\downarrow} + O^{(k)}_{\uparrow}\right).
\end{equation}
Finally, outputs from all groups are concatenated and projected with a $1\times1$ convolution, followed by normalization, activation, and a residual connection:

\begin{equation}
\label{eq:moss_out}
Y = X + \delta\!\left(\mathrm{Norm}\!\left(\mathrm{Proj}\!\left(\mathrm{Concat}\!\left(O^{(1)}, \ldots, O^{(K)}\right)\right)\right)\right),
\end{equation}
where $\mathrm{Proj}(\cdot)$ is a $1\times1$ convolution, $\mathrm{Norm}(\cdot)$ denotes batch normalization (or group normalization), and $\delta(\cdot)$ is ReLU. Eq.~\eqref{eq:moss_out} preserves the input dimensionality and enables direct substitution into the U-Net backbone. The complete forward computation of the proposed memory-modulated state-space gated block is summarized in Algorithm~\ref{alg:mossgate}.

\section{Experimental Setup}
\label{sec:experiments}

\subsection{Datasets and Preprocessing}
\label{sec:datasets}

Experiments are conducted on the \textbf{ISIC 2017} \cite{li2018skin} and \textbf{ISIC 2018} \cite{tschandl2018ham10000} skin lesion segmentation benchmarks, consisting of RGB dermoscopic images with pixel-wise binary lesion masks. For ISIC 2017, we use the official split of $2{,}000$ training, $150$ validation, and $600$ test images. ISIC 2018 provides $2{,}594$ training images; following common practice, a subset is held out for validation and evaluation is performed on the official test set. All experiments adhere to standard splits to ensure fair comparison with prior work. Images are resized to a fixed resolution of $352 \times 352$. Segmentation masks are resized using nearest-neighbor interpolation and binarized to $\{0,1\}$. Pixel intensities are normalized using dataset-independent normalization. During training, we apply on-the-fly data augmentation including random horizontal and vertical flips, random $90^\circ$ rotations, affine transformations with bounded translation, scaling and rotation, and color jittering. All augmentations are applied identically to images and masks. During validation and testing, only resizing and normalization are used.

\textbf{Implementation Details}
Models are implemented in PyTorch and trained on two NVIDIA RTX 4090 GPUs. The backbone is a standard five-stage U-Net with symmetric skip connections, where selected convolutional blocks in deep encoder stages are replaced by MoSSGate. Unless stated otherwise, MoSSGate is inserted only at the bottleneck stage to balance accuracy and efficiency. Batch normalization and ReLU activations are used throughout. Training is performed with mini-batches of $8$ images per GPU. Optimization uses AdamW with an initial learning rate of $1\times10^{-3}$ and weight decay $1\times10^{-5}$ for $80$ epochs, without learning-rate warmup. Mixed-precision training is employed for efficiency. The final model is selected based on the highest validation Dice score.

\subsection{Evaluation Metrics}
\label{sec:evaluation_metrics}
Segmentation performance is evaluated using the Dice Similarity Coefficient (DSC), Intersection over Union (IoU), and pixel-wise Accuracy (Acc), which are standard metrics for binary medical image segmentation. All metrics are computed on binarized prediction masks and averaged over the evaluation dataset. Let $P \in \mathbb{R}^{B \times 1 \times H \times W}$ denote the predicted logits and $Y \in \{0,1\}^{B \times 1 \times H \times W}$ the corresponding ground-truth masks. During evaluation, logits are converted to probabilities using the sigmoid function and thresholded at $0.5$ to obtain binary predictions:
\begin{equation}
\label{eq:threshold}
\hat{Y} = \mathbb{I}\bigl(\sigma(P) > 0.5\bigr),
\end{equation}
where $\mathbb{I}(\cdot)$ denotes the indicator function. The Dice Similarity Coefficient is defined as
\begin{equation}
\label{eq:dice_metric}
\mathrm{DSC}(\hat{Y}, Y) = \frac{2\,|\hat{Y} \cap Y| + \epsilon}{|\hat{Y}| + |Y| + \epsilon},
\end{equation}
where $|\cdot|$ denotes summation over spatial dimensions and $\epsilon$ is a small constant for numerical stability. The Intersection over Union is defined as
\begin{equation}
\label{eq:iou_metric}
\mathrm{IoU}(\hat{Y}, Y) = \frac{|\hat{Y} \cap Y| + \epsilon}{|\hat{Y} \cup Y| + \epsilon}.
\end{equation}
Pixel-wise Accuracy measures the proportion of correctly classified pixels over the entire image and is defined as
\begin{equation}
\label{eq:accuracy_metric}
\mathrm{Acc}(\hat{Y}, Y) = \frac{|\hat{Y} = Y|}{H \times W},
\end{equation}
where $|\hat{Y} = Y|$ counts the number of pixels for which the predicted label matches the ground truth. All metrics are computed per sample and then averaged across the dataset. The same preprocessing steps, thresholding strategy, and metric definitions are applied to all methods to ensure consistent and fair evaluation.

\begin{table*}[!ht]
\centering
\caption{Quantitative comparison on the ISIC2017 and ISIC2018 skin lesion segmentation benchmarks. Higher mIoU, Dice (DSC), and Accuracy (Acc) indicate better performance. FLOPs and parameter counts reflect computational and model complexity, respectively.}
\label{tab:isic_full}
\resizebox{\textwidth}{!}{
\begin{tabular}{l c c c c c c c c}
\hline
\multirow{2}{*}{Methods} & \multirow{2}{*}{FLOPs (G)} & \multirow{2}{*}{Params (M)} 
& \multicolumn{3}{c}{ISIC2017} & \multicolumn{3}{c}{ISIC2018} \\
\cline{4-9}
 &  &  & mIoU(\%)$\uparrow$ & DSC(\%)$\uparrow$ & Acc(\%)$\uparrow$
 & mIoU(\%)$\uparrow$ & DSC(\%)$\uparrow$ & Acc(\%)$\uparrow$ \\
\hline
UNet \cite{ronneberger2015u}            
& 14.101 & 7.853 
& 81.4$\pm$0.37 & 89.7$\pm$0.24 & 90.8$\pm$0.41 
& 80.1$\pm$0.33 & 88.9$\pm$0.28 & 89.7$\pm$0.45 \\

Att-UNet \cite{oktay2018attention}         
& 4.565  & 8.561 
& 83.0$\pm$0.22 & 90.8$\pm$0.35 & 92.1$\pm$0.18 
& 81.6$\pm$0.29 & 90.1$\pm$0.31 & 93.2$\pm$0.27 \\

SwinUNet \cite{cao2022swin}             
& 0.690  & 3.042 
& 79.7$\pm$0.46 & 88.6$\pm$0.39 & 93.5$\pm$0.21 
& 77.3$\pm$0.44 & 87.1$\pm$0.36 & 90.0$\pm$0.32 \\

UCM-Net \cite{yuan2024ucm}                
& \textbf{0.045} & 0.045 
& 83.2$\pm$0.25 & 90.9$\pm$0.19 & 95.0$\pm$0.14 
& 81.9$\pm$0.23 & 90.2$\pm$0.26 & 92.4$\pm$0.17 \\

U-KAN \cite{li2025u}                  
& 6.889  & 9.385 
& 82.4$\pm$0.34 & 90.4$\pm$0.28 & 94.6$\pm$0.16 
& 81.5$\pm$0.31 & 89.9$\pm$0.22 & 93.1$\pm$0.29 \\

Rolling-UNet \cite{liu2024rolling}     
& 8.282  & 7.096 
& 83.5$\pm$0.27 & 89.7$\pm$0.42 & 95.9$\pm$0.11 
& 83.9$\pm$0.24 & 90.1$\pm$0.37 & 94.3$\pm$0.20 \\

CMUNeXt \cite{tang2024cmunext}                
& 7.418  & 3.149 
& 85.0$\pm$0.18 & 90.8$\pm$0.23 & 95.4$\pm$0.15 
& 84.1$\pm$0.21 & 90.3$\pm$0.34 & 94.8$\pm$0.19 \\

\hline
\textbf{Ours}               
& 0.701  & 5.691 
& \textbf{86.3$\pm$0.12} & \textbf{92.6$\pm$0.17} & \textbf{96.6$\pm$0.09} 
& \textbf{85.9$\pm$0.14} & \textbf{90.6$\pm$0.16} & \textbf{95.4$\pm$0.13} \\

\hline
\end{tabular}
}
\end{table*}

\vspace{-16pt}
\section{Results and Analysis}
\label{sec:Results}
\subsection{Performance comparison with State-of-the-Art}
The quantitative results on ISIC2017, Table~\ref{tab:isic_full}, indicate that the proposed model attains the highest scores across all three metrics, reaching an mIoU of 86.3\%, DSC of 92.6\%, and accuracy of 96.6\%. This corresponds to an mIoU margin of around 1.3\% over the strongest CNN-based competitor CMUNeXt (85.0\%) and approximately 2.8-4.9\% over earlier UNet-style baselines such as Att-UNet (83.0\%) and UNet (81.4\%). Similar trends appear for DSC, where the gain is about 1.7\% over CMUNeXt (90.8\%) and around 1.8-2.9\% over Att-UNet and UNet. Compared with lightweight models, the mIoU exceeds UCM-Net by roughly 3.1\% and SwinUNet by about 6.6\%, while accuracy improves by around 1.6\% and 3.1\%, respectively. Relative to more complex recent designs such as Rolling-UNet and U-KAN, the mIoU increases by approximately 2.8\% and 3.9\%, and the DSC improves by about 2.9\% and 2.2\%. These differences indicate a consistent numerical margin over both heavy and lightweight alternatives while using fewer FLOPs than most CNN-based competitors.

For ISIC2018, a similar ordering is observed, with the proposed method achieving 85.9\% mIoU, 90.6\% DSC, and 95.4\% accuracy. The mIoU exceeds the closest CNN-based method CMUNeXt (84.1\%) by around 1.8\% and Rolling-UNet (83.9\%) by about 2.0\%, while remaining approximately 4.3-5.8\% above Att-UNet (81.6\%) and UNet (80.1\%). Against lightweight UCM-Net (81.9\%) and transformer-based SwinUNet (77.3\%), the mIoU margin increases to roughly 4.0\% and 8.6\%, respectively. For DSC, the gain over CMUNeXt is modest at about 0.3\%, but widens to around 0.5\% over Rolling-UNet and 2.5-3.5\% over classical baselines. Pixel accuracy improves by approximately 0.6\% over CMUNeXt and 1.1\% over Rolling-UNet, and by around 2.2-5.7\% compared with Att-UNet, UNet, and SwinUNet. The persistence of 1-2\% margins over recent strong CNN variants and larger 4-9\% margins over lightweight or transformer-heavy models suggests stable cross-dataset behavior.

\begin{figure}[!ht]
    \centering
    \includegraphics[width=1\linewidth]{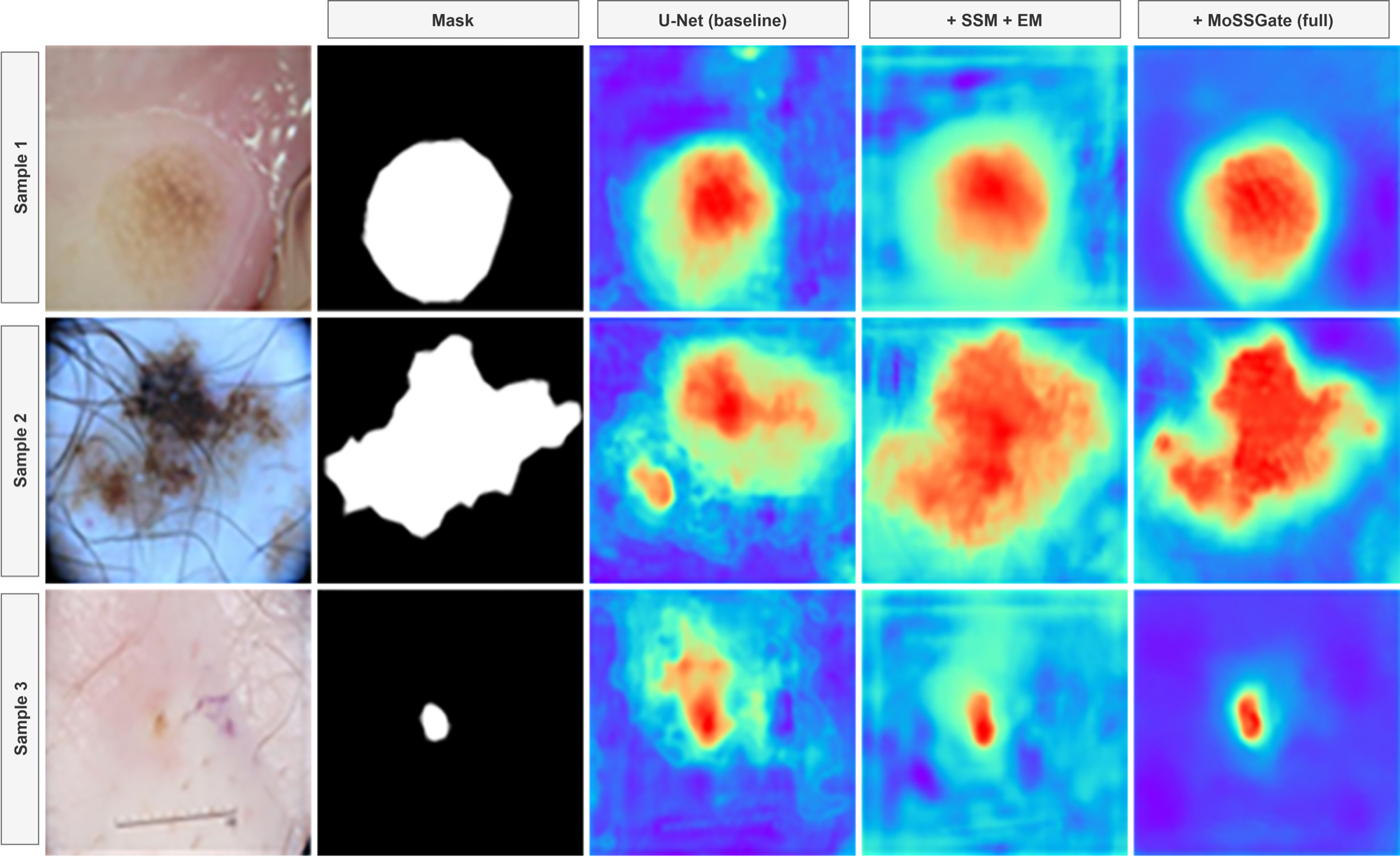}
    \caption{Qualitative heatmap comparison of intermediate feature responses for the baseline, partial, and full MoSSGate variants, showing progressively more concentrated and lesion-aligned activations as additional components are enabled.}
    \label{fig:heatmaps}
\end{figure}

\begin{figure*}[!ht]
    \centering
    \includegraphics[width=1\linewidth]{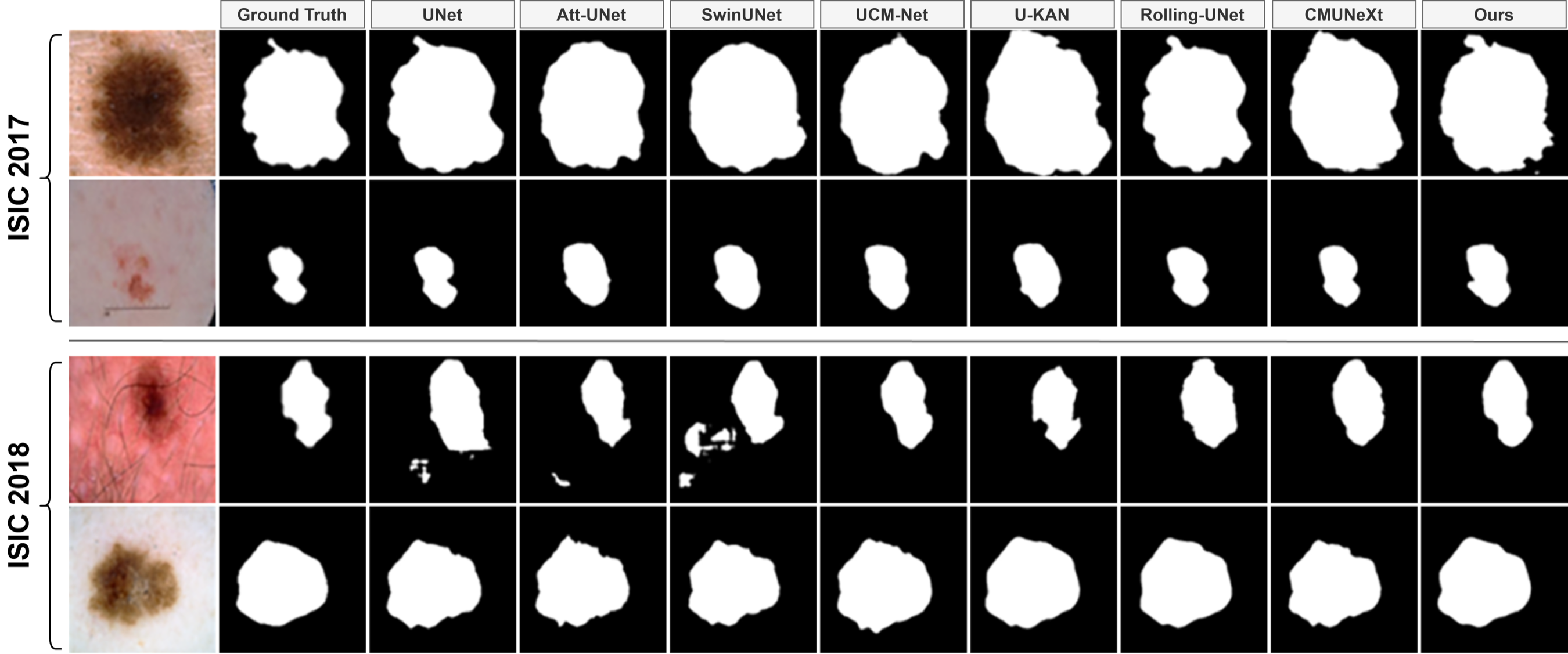}
    \caption{Qualitative comparison of skin lesion segmentation on representative ISIC2017 and ISIC2018 images. Compared methods show over-segmentation, fragmentation, or missed thin structures, while the proposed method better preserves fine boundaries and overall lesion shape.}
    \label{fig:qualitative_skin}
\end{figure*}

Qualitative comparisons reveal clear differences in boundary accuracy and region completeness across methods. CNN-based models such as U-Net and Att-UNet tend to produce overly smooth and slightly expanded contours, causing mild over-segmentation near irregular borders. SwinUNet often yields fragmented predictions with spurious false positives in challenging cases. Lightweight models like UCM-Net and U-KAN generate compact masks but may miss thin peripheral structures. Rolling-UNet and CMUNeXt improve global shape consistency, yet still exhibit minor boundary leakage or under-coverage in low-contrast areas. In contrast, as illustrated in Fig.~\ref{fig:qualitative_skin}, the proposed method more closely follows the ground-truth contours, preserving fine boundary details while suppressing isolated false detections, consistent with the superior overlap scores reported in Table~\ref{tab:isic_full}.

\subsection{Ablation Study}
\label{sec:ablation}
\begin{figure}[!ht]
    \centering
    \includegraphics[width=1\linewidth]{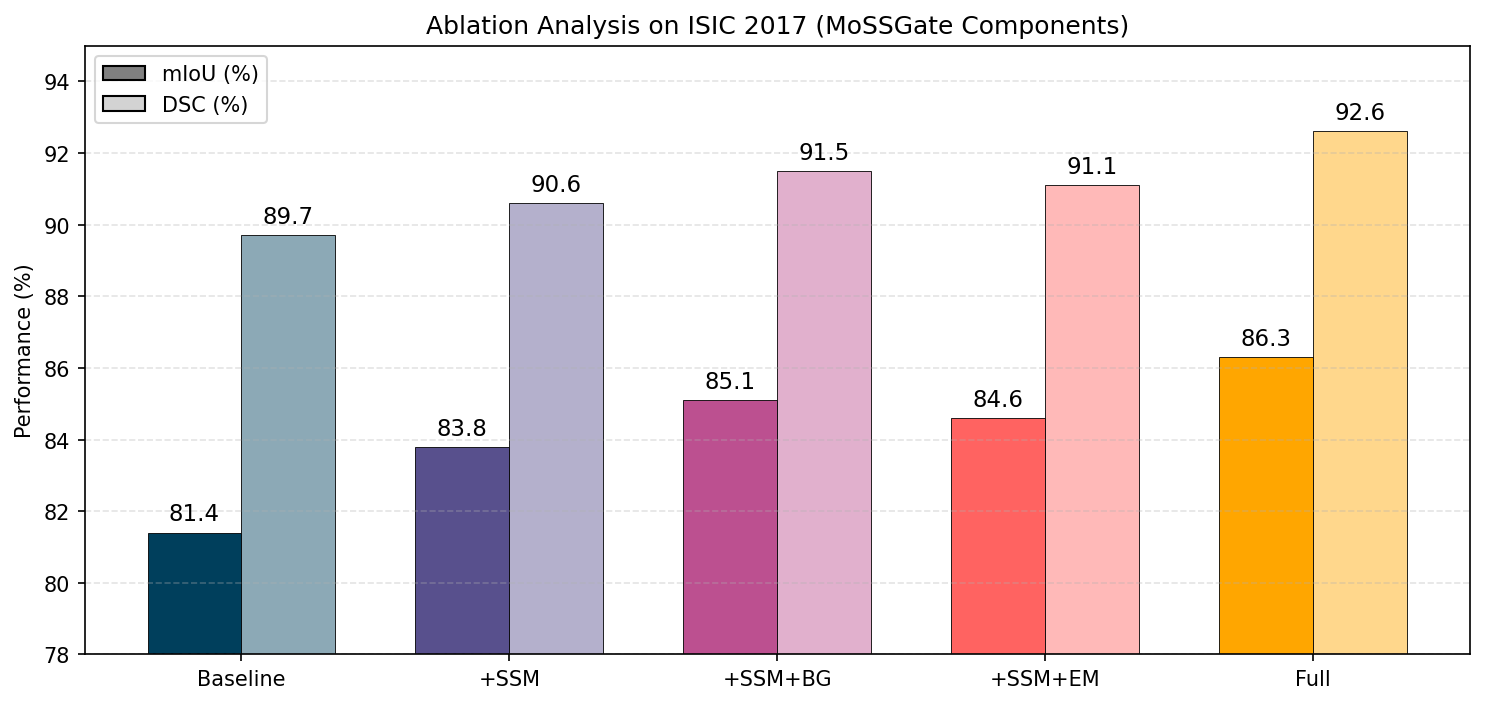}
    \caption{Ablation study of MoSSGate components on ISIC2017. Incrementally enabling state-space modeling (SSM), boundary gating (BG), and external memory modulation (EM) yields consistent improvements in mIoU and Dice, with the full configuration achieving the best performance.}
    \label{fig:ablation_isic2017}
\end{figure}

To quantify the individual and joint contributions of each component, we conduct an ablation study on ISIC 2017 by progressively enabling parallel 2D state-space modeling (SSM), the boundary gate (BG), and the external memory modulator (EM) on top of a U-Net baseline. Quantitative results are summarized in Table~\ref{tab:ablation_components}. The baseline achieves 81.4\% mIoU and 89.7\% DSC. Enabling SSM alone improves performance to 83.8\% mIoU and 90.6\% DSC, demonstrating the benefit of efficient long-range spatial modeling. Adding BG further boosts the scores to 85.1\% mIoU and 91.5\% DSC, indicating that boundary-aware spatial gating enhances localization. Combining SSM with EM instead yields 84.6\% mIoU and 91.1\% DSC, highlighting the role of sample-adaptive contextual modulation. The full MoSSGate configuration integrating SSM, BG, and EM achieves the best results of 86.3\% mIoU and 92.6\% DSC, outperforming the baseline by +4.9 and +2.9 percentage points, respectively, and exceeding the strongest two-component variant (SSM+BG) by +1.2 mIoU and +1.1 DSC. As illustrated in Fig.~\ref{fig:ablation_isic2017}, progressively adding SSM, BG, and EM consistently improves both metrics, confirming that the three modules provide complementary gains. Feature visualizations in Fig.~\ref{fig:heatmaps} further support this trend. The baseline U-Net exhibits diffuse and spatially scattered activations with background leakage, while introducing SSM and EM concentrates responses over the lesion area. The full MoSSGate produces the most focused and boundary-aligned activations, indicating improved boundary sensitivity and context-aware feature integration. All reported main results use the full MoSSGate configuration (SSM+BG+EM) applied at the bottleneck stage.
\begin{table}[!ht]
\centering
\caption{Ablation study of MoSSGate components on ISIC 2017. BG denotes the boundary gate, EM denotes the external memory modulator, and SSM denotes the parallel 2D state-space modeling.}
\label{tab:ablation_components}
\begin{tabular}{l c c c c c}
\hline
Configuration & BG & EM & SSM & mIoU (\%)$\uparrow$ & DSC (\%)$\uparrow$ \\
\hline
U-Net (baseline)        & -- & -- & -- & 81.4 & 89.7 \\
+ SSM only              & -- & -- & \checkmark & 83.8 & 90.6 \\
+ SSM + BG              & \checkmark & -- & \checkmark & 85.1 & 91.5 \\
+ SSM + EM              & -- & \checkmark & \checkmark & 84.6 & 91.1 \\
\textbf{MoSSGate (full)}& \checkmark & \checkmark & \checkmark & \textbf{86.3} & \textbf{92.6} \\
\hline
\end{tabular}
\end{table}

\section{Conclusion}
\label{sec:conclusion}

We presented MoSSGate, a memory-modulated state-space gating module for efficient and boundary-aware skin lesion segmentation within a standard U-Net framework. By combining boundary-driven spatial gating, external memory modulation, and parallel 2D state-space modeling, the proposed module enables controlled long-range context propagation while preserving fine lesion boundaries. MoSSGate is a plug-and-play component that can be inserted into deep U-Net stages without altering the overall architecture. Experiments on the ISIC 2017 and ISIC 2018 benchmarks demonstrate state-of-the-art accuracy with a strong accuracy-efficiency trade-off compared to classical CNN, lightweight, and transformer models. Ablation and qualitative analyses confirm that boundary gating, memory modulation, and state-space modeling provide complementary benefits, yielding sharper boundaries and reduced background leakage when combined. These results highlight the potential of state-space models as an efficient alternative to attention for dense medical image segmentation when coupled with spatial control and adaptive dynamics. Although evaluated on skin lesion segmentation, the proposed design is general and can extend to other medical imaging tasks that require efficient global context modeling and precise boundary localization.

%
%
\bibliographystyle{splncs04}
\bibliography{references}

\begin{thebibliography}{10}
\providecommand{\url}[1]{\texttt{#1}}
\providecommand{\urlprefix}{URL }
\providecommand{\doi}[1]{https://doi.org/#1}

\bibitem{abbas2025gradient}
Abbas, J., Abbas, S., Liu, L.: Gradient-guided causal attention mechanism for
  interpretable skin lesion classification. In: Chinese Conference on Pattern
  Recognition and Computer Vision (PRCV). pp. 336--349. Springer (2025)

\bibitem{abbas2025dualattendmed}
Abbas, J., Soomro, D.B., Huang, S., Liu, L.: Dualattendmed: A coarse-to-fine
  dual-stage attention framework for interpretable disease localization and
  classification. Expert Systems with Applications p. 130886 (2025)

\bibitem{GeGLUNet}
Abdun~Noor, A.F.M., Ahasan, M.I., Khan, M.A., Yang, G.: Geglunet: Structural
  retinal vessel segmentation via attention-gated geglu and contrastive
  supervision. In: Pattern Recognition and Computer Vision (PRCV). pp.
  494--507. Springer (2025)

\bibitem{ates2023dual}
Ates, G.C., Mohan, P., Celik, E.: Dual cross-attention for medical image
  segmentation. Engineering Applications of Artificial Intelligence
  \textbf{126},  107139 (2023)

\bibitem{badar2025transformer}
Badar, D., Abbas, J., Alsini, R., Abbas, T., ChengLiang, W., Daud, A.:
  Transformer attention fusion for fine grained medical image classification.
  Scientific Reports  \textbf{15}(1),  20655 (2025)

\bibitem{cao2022swin}
Cao, H., Wang, Y., Chen, J., Jiang, D., Zhang, X., Tian, Q., Wang, M.:
  Swin-unet: Unet-like pure transformer for medical image segmentation. In:
  European conference on computer vision. pp. 205--218. Springer (2022)

\bibitem{chen2023transattunet}
Chen, B., Liu, Y., Zhang, Z., Lu, G., Kong, A.W.K.: Transattunet: Multi-level
  attention-guided u-net with transformer for medical image segmentation. IEEE
  Transactions on Emerging Topics in Computational Intelligence  \textbf{8}(1),
   55--68 (2023)

\bibitem{chen2021transunet}
Chen, J., Lu, Y., Yu, Q., Luo, X., Adeli, E., Wang, Y., Lu, L., Yuille, A.L.,
  Zhou, Y.: Transunet: Transformers make strong encoders for medical image
  segmentation. arXiv preprint arXiv:2102.04306  (2021)

\bibitem{gu2024mamba}
Gu, A., Dao, T.: Mamba: Linear-time sequence modeling with selective state
  spaces. In: First conference on language modeling (2024)

\bibitem{gu2021efficiently}
Gu, A., Goel, K., R{\'e}, C.: Efficiently modeling long sequences with
  structured state spaces. arXiv preprint arXiv:2111.00396  (2021)

\bibitem{he2023h2former}
He, A., Wang, K., Li, T., Du, C., Xia, S., Fu, H.: H2former: An efficient
  hierarchical hybrid transformer for medical image segmentation. IEEE
  Transactions on Medical Imaging  \textbf{42}(9),  2763--2775 (2023)

\bibitem{huang2025lmfa}
Huang, Q., Pan, F., Xu, Z., Wang, J.: Lmfa-unet: A lightweight
  multi-granularity feature adaptation unet for skin lesion segmentation. In:
  2025 International Joint Conference on Neural Networks (IJCNN). pp.~1--8.
  IEEE (2025)

\bibitem{khan2025medfusion}
Khan, M.Y., Abbas, J., Soomro, D.B., Yin, Y.: Medfusion-xai: An attention
  refinement and gated fusion explainable swin transformer for medical image
  classification. In: 2025 5th International Conference on Digital Futures and
  Transformative Technologies (ICoDT2). pp.~1--6. IEEE (2025)

\bibitem{lee2020structure}
Lee, H.J., Kim, J.U., Lee, S., Kim, H.G., Ro, Y.M.: Structure boundary
  preserving segmentation for medical image with ambiguous boundary. In:
  Proceedings of the IEEE/CVF conference on computer vision and pattern
  recognition. pp. 4817--4826 (2020)

\bibitem{li2025u}
Li, C., Liu, X., Li, W., Wang, C., Liu, H., Liu, Y., Chen, Z., Yuan, Y.: U-kan
  makes strong backbone for medical image segmentation and generation. In:
  Proceedings of the AAAI Conference on Artificial Intelligence. vol.~39, pp.
  4652--4660 (2025)

\bibitem{li2025vmc}
Li, H., Song, K., Zhang, P., Yu, H., Bian, H.: Vmc-unet: A u-shaped structure
  combining mamba and cnn for medical image segmentation. In: 2025
  International Joint Conference on Neural Networks (IJCNN). pp.~1--8. IEEE
  (2025)

\bibitem{li2025edgrnet}
Li, W., Wang, S., Chen, N., Lu, R., Shi, X.: Edgrnet: An error detection and
  guided refinement network for skin lesion segmentation. In: 2025
  International Joint Conference on Neural Networks (IJCNN). pp.~1--8. IEEE
  (2025)

\bibitem{li2018skin}
Li, Y., Shen, L.: Skin lesion analysis towards melanoma detection using deep
  learning network. Sensors  \textbf{18}(2), ~556 (2018)

\bibitem{lin2023rethinking}
Lin, Y., Zhang, D., Fang, X., Chen, Y., Cheng, K.T., Chen, H.: Rethinking
  boundary detection in deep learning models for medical image segmentation.
  In: International conference on information processing in medical imaging.
  pp. 730--742. Springer (2023)

\bibitem{liu2024vmamba}
Liu, Y., Tian, Y., Zhao, Y., Yu, H., Xie, L., Wang, Y., Ye, Q., Jiao, J., Liu,
  Y.: Vmamba: Visual state space model. Advances in neural information
  processing systems  \textbf{37},  103031--103063 (2024)

\bibitem{liu2024rolling}
Liu, Y., Zhu, H., Liu, M., Yu, H., Chen, Z., Gao, J.: Rolling-unet:
  Revitalizing mlp’s ability to efficiently extract long-distance
  dependencies for medical image segmentation. In: Proceedings of the AAAI
  conference on artificial intelligence. vol.~38, pp. 3819--3827 (2024)

\bibitem{mehta2022long}
Mehta, H., Gupta, A., Cutkosky, A., Neyshabur, B.: Long range language modeling
  via gated state spaces. arXiv preprint arXiv:2206.13947  (2022)

\bibitem{oktay2018attention}
Oktay, O., Schlemper, J., Folgoc, L.L., Lee, M., Heinrich, M., Misawa, K.,
  Mori, K., McDonagh, S., Hammerla, N.Y., Kainz, B., et~al.: Attention u-net:
  Learning where to look for the pancreas. arXiv preprint arXiv:1804.03999
  (2018)

\bibitem{qiu2021slimconv}
Qiu, J., Chen, C., Liu, S., Zhang, H.Y., Zeng, B.: Slimconv: Reducing channel
  redundancy in convolutional neural networks by features recombining. IEEE
  Transactions on Image Processing  \textbf{30},  6434--6445 (2021)

\bibitem{ronneberger2015u}
Ronneberger, O., Fischer, P., Brox, T.: U-net: Convolutional networks for
  biomedical image segmentation. In: International Conference on Medical image
  computing and computer-assisted intervention. pp. 234--241. Springer (2015)

\bibitem{ruan2022malunet}
Ruan, J., Xiang, S., Xie, M., Liu, T., Fu, Y.: Malunet: A multi-attention and
  light-weight unet for skin lesion segmentation. In: 2022 IEEE International
  Conference on Bioinformatics and Biomedicine (BIBM). pp. 1150--1156. IEEE
  (2022)

\bibitem{ruan2023ege}
Ruan, J., Xie, M., Gao, J., Liu, T., Fu, Y.: Ege-unet: an efficient group
  enhanced unet for skin lesion segmentation. In: International conference on
  medical image computing and computer-assisted intervention. pp. 481--490.
  Springer (2023)

\bibitem{tang2024cmunext}
Tang, F., Ding, J., Quan, Q., Wang, L., Ning, C., Zhou, S.K.: Cmunext: An
  efficient medical image segmentation network based on large kernel and skip
  fusion. In: 2024 IEEE International Symposium on Biomedical Imaging (ISBI).
  pp.~1--5. IEEE (2024)

\bibitem{tschandl2018ham10000}
Tschandl, P., Rosendahl, C., Kittler, H.: The ham10000 dataset, a large
  collection of multi-source dermatoscopic images of common pigmented skin
  lesions. Scientific data  \textbf{5}(1), ~1--9 (2018)

\bibitem{ashish2017attention}
Vaswani, A., Shazeer, N., Parmar, N., Uszkoreit, J., Jones, L., Gomez, A.N.,
  Kaiser, {\L}., Polosukhin, I.: Attention is all you need. Advances in neural
  information processing systems  \textbf{30} (2017)

\bibitem{wang2022uctransnet}
Wang, H., Cao, P., Wang, J., Zaiane, O.R.: Uctransnet: rethinking the skip
  connections in u-net from a channel-wise perspective with transformer. In:
  Proceedings of the AAAI conference on artificial intelligence. vol.~36, pp.
  2441--2449 (2022)

\bibitem{wang2021boundary}
Wang, J., Wei, L., Wang, L., Zhou, Q., Zhu, L., Qin, J.: Boundary-aware
  transformers for skin lesion segmentation. In: International conference on
  medical image computing and computer-assisted intervention. pp. 206--216.
  Springer (2021)

\bibitem{wang2023selective}
Wang, J., Zhu, W., Wang, P., Yu, X., Liu, L., Omar, M., Hamid, R.: Selective
  structured state-spaces for long-form video understanding. In: Proceedings of
  the IEEE/CVF Conference on Computer Vision and Pattern Recognition. pp.
  6387--6397 (2023)

\bibitem{wang2022eanet}
Wang, K., Zhang, X., Zhang, X., Lu, Y., Huang, S., Yang, D.: Eanet: Iterative
  edge attention network for medical image segmentation. Pattern Recognition
  \textbf{127},  108636 (2022)

\bibitem{wu2022fat}
Wu, H., Chen, S., Chen, G., Wang, W., Lei, B., Wen, Z.: Fat-net: Feature
  adaptive transformers for automated skin lesion segmentation. Medical image
  analysis  \textbf{76},  102327 (2022)

\bibitem{wu2024mhorunet}
Wu, R., Liang, P., Huang, X., Shi, L., Gu, Y., Zhu, H., Chang, Q.: Mhorunet:
  High-order spatial interaction unet for skin lesion segmentation. Biomedical
  Signal Processing and Control  \textbf{88},  105517 (2024)

\bibitem{wu2024hsh}
Wu, R., Lv, H., Liang, P., Cui, X., Chang, Q., Huang, X.: Hsh-unet: Hybrid
  selective high order interactive u-shaped model for automated skin lesion
  segmentation. Computers in Biology and Medicine  \textbf{168},  107798 (2024)

\bibitem{yu2022s2}
Yu, T., Li, X., Cai, Y., Sun, M., Li, P.: S2-mlp: Spatial-shift mlp
  architecture for vision. In: Proceedings of the IEEE/CVF winter conference on
  applications of computer vision. pp. 297--306 (2022)

\bibitem{yuan2024ucm}
Yuan, C., Zhao, D., Agaian, S.S.: Ucm-net: A lightweight and efficient solution
  for skin lesion segmentation using mlp and cnn. Biomedical Signal Processing
  and Control  \textbf{96},  106573 (2024)

\bibitem{zhao2023mms}
Zhao, C., Lv, W., Zhang, X., Yu, Z., Wang, S.: Mms-net: multi-level multi-scale
  feature extraction network for medical image segmentation. Biomedical Signal
  Processing and Control  \textbf{86},  105330 (2023)

\bibitem{zhao2023m}
Zhao, X., Jia, H., Pang, Y., Lv, L., Tian, F., Zhang, L.: M$^2$snet:
  Multi-scale in multi-scale subtraction network for medical image
  segmentation. arXiv preprint  (2023)

\end{thebibliography}

\end{document}